\documentclass[conference]{IEEEtran}
\IEEEoverridecommandlockouts
\usepackage{cite}
\usepackage{amsmath,amssymb,amsfonts}
\usepackage{algorithmic}
\usepackage{graphicx}
\usepackage{textcomp}
\usepackage{xcolor}
\usepackage[colorlinks=true, allcolors=blue]{hyperref}
\usepackage{float}
\usepackage{tabularx}
\usepackage[english]{babel}
\usepackage{enumitem}
\usepackage{tabularx}
\usepackage{afterpage}
\usepackage{kantlipsum}
\usepackage{dblfloatfix} 
\usepackage{caption}
\usepackage{subcaption}
\usepackage{balance}
\renewcommand{\arraystretch}{1.2}
\newlist{steps}{enumerate}{1}
\setlist[steps, 1]{label = Step \arabic*:}
\def\BibTeX{{\rm B\kern-.05em{\sc i\kern-.025em b}\kern-.08em
    T\kern-.1667em\lower.7ex\hbox{E}\kern-.125emX}}
\begin{document}

\setlength{\textfloatsep}{2pt plus 1pt minus 2pt}
\setlength{\floatsep}{2pt plus 1pt minus 2pt}
\setlength{\intextsep}{2pt plus 1pt minus 2pt}

\setlength{\abovecaptionskip}{1pt}
\setlength{\belowcaptionskip}{1pt}

\setlength{\tabcolsep}{3pt}
\renewcommand{\arraystretch}{1}

\setlist[itemize]{topsep=2pt, itemsep=1pt, parsep=0pt, partopsep=0pt}
\setlist[enumerate]{topsep=2pt, itemsep=1pt, parsep=0pt, partopsep=0pt}

\setlength{\abovedisplayskip}{4pt}
\setlength{\belowdisplayskip}{4pt}
\setlength{\abovedisplayshortskip}{3pt}
\setlength{\belowdisplayshortskip}{3pt}

\title{Think Less, Label Better: Multi-Stage Domain-Grounded Synthetic Data Generation for Fine-Tuning Large Language Models in Telecommunications}
\author{
Chenhua Shi$^{*}$
\quad Gregor Macdonald$^{*}$
\quad Bhavika Jalli
\quad Wanlu Lei
\quad John Zou
\quad Mridul Jain
\quad Joji Philip 
\\[0.25em]
\texttt{Ericsson} \\[0.25em]
\thanks{$^{*}$Equal contribution}
}


\maketitle

\begin{abstract}
The success of large language models (LLMs) depends heavily on large-scale, high-quality instruction-following and reinforcement datasets. However, generating such data through human annotation is prohibitively time-consuming particularly for domain-specific tasks like telecom network troubleshooting, where accurate responses require deep technical expertise and contextual understanding. In this paper, we present a fully automated, retrieval-augmented pipeline for generating synthetic question-answer (QA) pairs grounded in structured domain knowledge. Our multi-stage framework integrates a retriever, base generator, and refinement model to synthesize and enhance QA pairs using documents retrieved from a domain-specific knowledge graph. To ensure data quality, we employ customized RAGAS-based scoring to filter low-quality samples, producing a high-quality dataset suitable for reinforcement fine-tuning (RFT). We demonstrate our approach in a real-world telecom scenario focused on radio access network (RAN) troubleshooting. The resulting pipeline generates complex, context-rich troubleshooting solution plans without human intervention. This work offers a scalable solution for building instruction and reinforcement datasets in specialized domains, significantly reducing dependence on manual labeling while maintaining high technical fidelity.
\end{abstract}


\section{Introduction}
Foundation models \cite{foundationmodels} have transformed natural language processing by demonstrating strong generalization across a wide range of tasks. However, these models often struggle in high-stakes, specialized domains—such as telecom, medicine, and law—where responses must be grounded in precise, domain-specific knowledge. Adapting LLMs to such environments typically requires large-scale, high-quality labeled data, particularly for supervised fine-tuning (SFT) and reinforcement fine-tuning (RFT). Unfortunately, producing these datasets through manual annotation is prohibitively expensive, time-consuming, and often infeasible due to the need for deep technical expertise.

While non-parametric methods such as retrieval-augmented generation (RAG)\cite{rag} \cite{ragsurvey} offer a lightweight way to inject external knowledge at inference time, they often fail to fully leverage a model’s parametric reasoning capabilities. Furthermore, RAG systems lack explicit control over the structure and quality of generated outputs, limiting their utility for creating consistent fine-tuning datasets.

To address these limitations, we propose a fully automated, multi-stage pipeline for synthetic QA data generation tailored to instruction tuning and RFT in specialized domains. Our approach eliminates the need for human labeling by orchestrating three LLM components: a retriever, a base generator, and a refinement model. The pipeline begins with the retrieval of domain-relevant context from a structured knowledge graph using HippoRAG\cite{hipporag}. A base generation model then synthesizes candidate QA pairs grounded in this context. Next, the refinement model leverages the top-k most relevant documents from HippoRAG to improve and summarize the generated answers. Finally, low-quality outputs are filtered using customized RAGAS \cite{ragas} scores, resulting in a high-fidelity dataset well-suited for RFT.

We demonstrate the proposed framework in the context of telecom network troubleshooting, a domain where accurate solution plans are critical. In this setting, inputs may take the form of engineer-specified queries—for example, how to troubleshoot specific alarms or resolve a network faults. The framework outputs detailed, step-by-step troubleshooting solution plans that integrate evidence from multiple sources, including alarm logs, configuration management, performance counters, etc. The resulting synthetic QA dataset not only reflects the domain specificity and technical rigor of expert-authored responses but also significantly accelerates the preparation of high-quality training data for RFT.

\section{Related Work}
Synthetic data has become a cornerstone in scaling LLM performance, as shown by recent models such as Mixtral-8x7B \cite{mixtral}, LLaMA 3.3 \cite{lama3.3}, and Qwen 3 \cite{qwen3}. These rely heavily on large volumes of synthetic instruction-following or conversational data generated from prompts, templates, or seed tasks. However, over-reliance on synthetic data risks model collapse, where repeated fine-tuning on low-diversity or self-generated distributions leads to mode-seeking behavior, reduced diversity, and degraded downstream performance \cite{modelcollapse}. Frameworks like BARE \cite{bare} mitigate this by introducing a base–refine pipeline that generates diverse data from only a few examples, avoiding extensive prompt engineering. While effective at improving diversity, BARE lacks structured grounding and can hallucinate—limitations that are especially problematic in technical domains requiring factual precision. Similarly, the Synthetic Data RL framework \cite{syntheticrl} combines knowledge-guided synthesis with reinforcement learning and adaptive curricula, but its reliance on instruct-tuned models constrains domain-specific reasoning and generalization. Such pipelines remain weak on tasks like root-cause analysis or multi-step troubleshooting, where correctness depends on expert knowledge and contextual retrieval beyond simple prompts.  

In telecommunications, recent works such as TelecomGPT \cite{telecomgpt} and TSLAM-Mini \cite{tslam-mini} adapt LLMs through public web sources, heavy data processing, and fine-tuning. While these approaches achieve domain alignment, their datasets are largely descriptive or digital twin based simulation \cite{tslam-mini} and do not tackle the harder challenge of constructing reasoning-intensive datasets for troubleshooting and fault analysis.  

Lastly, synthetic data quality is commonly evaluated through downstream benchmarks such as GSM8K, SWE-bench, TruthfulQA, etc, under the assumption that better data leads to better generalization \cite{bare,syntheticrl}. Although useful, such evaluations only indirectly reflect sample quality. Token- and embedding-based metrics like BLEU \cite{blue}, BERTScore \cite{bertscore}, and Sentence-BERT \cite{bertsentence} offer finer granularity but fall short for telecom domain requiring step-by-step procedural reasoning. To address this, we employ RAGAS to score QA pairs on faithfulness, relevance, and correctness relative to retrieved context, providing factual grounding and stepwise validity—criteria critical for evaluating synthetic troubleshooting plans and fault analysis in telecom networks.

\section{System Overview and Methods}

\begin{figure}[htbp!]
\centering
\includegraphics[width=0.85\linewidth]{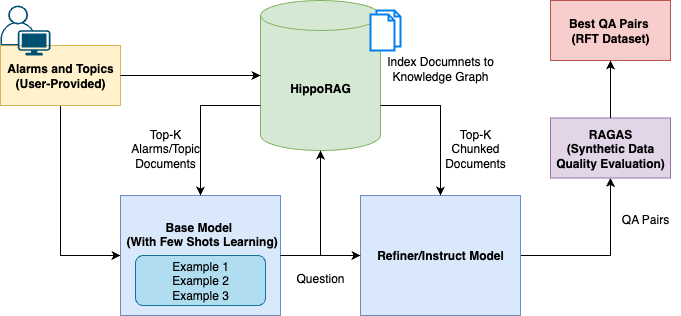}
\caption{\label{fig:architecture}The architecture diagram for Multi-Stage
Domain-Grounded Synthetic Data Generation for
Fine-Tuning LLMs.}
\end{figure}

We present a multi-stage synthetic QA generation pipeline (Figure \ref{fig:architecture}) designed to produce high-quality QA pairs for fine-tuning LLMs in domain-specific contexts such as telecom troubleshooting. Building on the BARE framework—which demonstrates the ability to generate diverse, high-quality synthetic datasets from open-source data—our approach extends it by incorporating external knowledge sources to enhance both diversity and domain grounding. To further ensure data fidelity, we introduce customized RAGAS metrics, including a telecom-specific specificity measure, to filter out low-quality outputs. The pipeline integrates multiple LLMs, each fulfilling a specialized role to promote question diversity, factual accuracy, procedural clarity, and strong contextual grounding.

The process begins with structured network fault alarms serving as prompts, along with topic documents that provide relevant background knowledge. A HippoRAG retriever module, powered by Qwen3-8B, identifies the top-3 relevant document chunks from the topic corpus using semantic similarity based on the built-knowledge graph. These retrieved chunks form the context for the initial generation of QA. The initial generation step uses a base model (Qwen-14B-Base) to synthesize raw QA pairs. This model is intentionally not instruct-tuned, allowing it to generate diverse and unbiased QA pairs. The outputs are then passed to an instruct-tuned model (Qwen3-8B/Qwen3-32B) with its `thinking' mode — i.e., explicit chain-of-thought decoding — disabled for efficiency. In this configuration, the model directly produces final answers without emitting intermediate reasoning tokens. This model is provided with the top-3 supporting document chunks retrieved when HippoRAG is queried by the question, and rewrites the QA pairs for improved coherence, factual grounding, and clarity. The refinement emphasizes structured reasoning, which is essential for generating troubleshooting plans that follow logical step-by-step procedures. To ensure quality, we apply a filtering step using custom RAGAS scores, again powered by Qwen3-8B. Each QA pair is scored on key dimensions such as response groundedness, response relevancy, specificity, and aspect critic. Finally, only QA pairs above pre-defined thresholds are retained to form the RFT dataset.

This pipeline provides a scalable, domain-adaptable method for generating procedurally correct, context-aware synthetic data, designed to capture procedural correctness and structured reasoning without relying on human annotation.

\subsection{Approach}

\subsubsection{External Knowledge-Guided Synthesis}

To generate domain-grounded synthetic data and minimize hallucinations, we incorporate external knowledge retrieval into the QA generation process. We leverage HippoRAG as our retrieval backbone, as it consistently outperformed standard RAG and Graph RAG in internal evaluations. HippoRAG enhances grounding quality by integrating LLM-extracted triples, phrase-level and passage-level knowledge graphs, and a personalized page-rank algorithm that prioritizes relevant context retrieval. We index a corpus of telecom troubleshooting documents —including fault alarms, performance counters, and configuration management— into HippoRAG. This external knowledge source serves two critical roles in our pipeline: 
\begin{enumerate}
    \item Ensure generated questions for different alarms and topics are contextually relevant and aligned with real operational content during the initial QA generation phase.
    \item Provide reliable, domain-specific grounding to support the generation of accurate, multi-step solution plans during the refinement phase.
\end{enumerate}

We utilize a set of domain-specific network fault alarms, topics, or questions $K$, as keys for identifying relevant context within an external passage library $L$. From $L$, we retrieve a set of semantically relevant passages $R$ from passage retriever $P$, which serve as grounding evidence for subsequent QA generation in both the base and refinement phases \eqref{retriever}:

\vspace{-0.5em}
\begin{equation}
R = P(K, L) \label{retriever}
\end{equation}

\subsubsection{Customized RAGAS Evaluation on Domain-Grounded Synthetic Data}

To evaluate the quality of domain-grounded synthetic QA pairs, we extend standard RAGAS evaluation with metrics tailored to the requirements of telecom troubleshooting. While conventional measures such as ResponseRelevancy and ResponseGroundedness capture general alignment with retrieved context, they fall short in domains that demand precise use of alarms, counters, configurations, and ordered troubleshooting steps. To address this, we introduce customized metrics and thresholds to ensure domain specificity, factual accuracy, and procedural consistency. In particular, Tele-Specificity verifies that domain-specific terms in both questions and answers are supported by the retrieved context, mitigating hallucination, while AspectCritic filters out unanswerable cases. Together, these metrics ensure that the retained QA pairs are contextually grounded, technically specific, and operationally reliable—an essential requirement for high-stakes domains such as telecom troubleshooting.

\subsubsection{ResponseRelevancy}
Response Relevancy measures how directly the generated answer addresses the given question. It penalizes responses that are incomplete, redundant, or contain unnecessary information—regardless of factual accuracy. The score ranges from 0 to 1, with 1 indicating a highly relevant answer. To compute this, we prompt an LLM to generate $n$ alternative questions based on the answer $a_s(q)$ and embed them using Qwen3-8B Embedding model. We then calculate the cosine similarity between each generated question $q_i$ and the original one $q$. The average similarity defines the final relevance score $RS$ \eqref{RS}, reflecting how well the answer remains focused and aligned with the user's intent.

\begin{equation}
RS = \sum_{i=1}^{n} cos(q, q_i) \label{RS}
\end{equation}

\subsubsection{ResponseGroundedness} Response Groundedness measures how well the generated answer is supported by the retrieved context. Using an LLM-guided evaluation, we prompt the model with templated instructions to classify the answer's grounding into three levels:
\begin{itemize}
\item Score 0: Not grounded — the response is unsupported or irrelevant to the context
\item Score 0.5: Partially grounded — the response is somewhat supported but lacks completeness
\item Score 1: Fully grounded — the response is fully supported by the provided evidence
\end{itemize}

\subsubsection{Tele-Specificity}
Tele-Specificity evaluates whether both the question and the response contain domain-grounded terminology—such as alarms, performance counters, and configurations—that is also present in the retrieved context.

For question specificity, the LLM is instructed to return:
\begin{itemize}
\item Score 0: if any required term is missing from the retrieved context or if no such terms appear in the question
\item Score 1: if all terms in the question are present in the retrieved context
\end{itemize}
For response specificity, the LLM is instructed to return:
\begin{itemize}
\item Score 0: if any required term is missing from the retrieved context or if no such terms appear in the response
\item Score 1: if all terms are present in the retrieved context and at least one alarm or performance counter is mentioned in the response
\end{itemize}

Few-shot examples are provided to help the LLM assign accurate scores. The final specificity score is computed as the average of the question and response scores, with a value of 1 only if both are fully matched. This guards against hallucination and ensures that generated outputs reflect real, actionable telecom troubleshooting scenarios.

\subsubsection{AspectCritic}
AspectCritic detects whether a given question can be reasonably answered from the retrieved context. Using an LLM-based evaluator, the metric returns:
\begin{itemize}
\item Score 0: if the question cannot be answered based on the provided context 
\item Score 1: if the context contains sufficient information to answer the question
\end{itemize}

This check helps identify gaps in the retrieval process and prevents the generation of speculative or hallucinated responses, ensuring that all synthetic QA pairs remain grounded in verifiable telecom troubleshooting knowledge.

\section{Experiments}
We conduct our experiments on a cluster equipped with seven NVIDIA RTX A6000 GPUs, each with 48 GB of VRAM. Model serving is managed using vLLM \cite{vllm} to efficiently host multiple large models. In our setup, we use the Qwen3 family of LLMs. Qwen3-14B-Base serves as the base generator on two GPUs, Qwen3-8B/Qwen3-32B acts as the instruct-tuned model on another two GPUs, and Qwen3-8B is used for RAGAS-based evaluation on a single GPU. The remaining two GPUs are allocated to HippoRAG, utilizing Qwen3-8B as the retrieval model. We benchmark the hybrid model against both the base-only and instruct-tuned-only models across two datasets, demonstrating why it offers the most effective option for curating RFT datasets.

\subsection{Seed Data Curation}
We begin with seed data preparation to provide few-shot examples for both the base stage and refinement stage of domain-grounded synthetic data generation. We manually create 50 questions derived from troubleshooting documents covering diverse topics and alarms related to power systems in the telecom domain. For each question, we use HippoRAG in combination with  LLMs (GPT-4o, LLaMA-3.3-70B-Instruct, and DeepSeek-R1-70B) to generate detailed stepwise solution plans. These QA pairs are then verified and refined by subject matter experts (SMEs) to ensure technical accuracy and completeness. Each finalized seed sample consists of three components: the question, the answer, and the top-3 relevant context chunks retrieved by HippoRAG. Although the limited number of expert-curated seed QA pairs may constrain question diversity, they are used only as few-shot exemplars to steer the base model toward diverse QA generation, and retrieval-grounded RAGAS metrics with strict answerability and groundedness filtering ensure the robustness and quality of the resulting synthetic dataset.

\subsection{Domain-Grounded Synthetic Data Generation}

\begin{figure}[htbp!]
\centering
\includegraphics[width=1\linewidth]{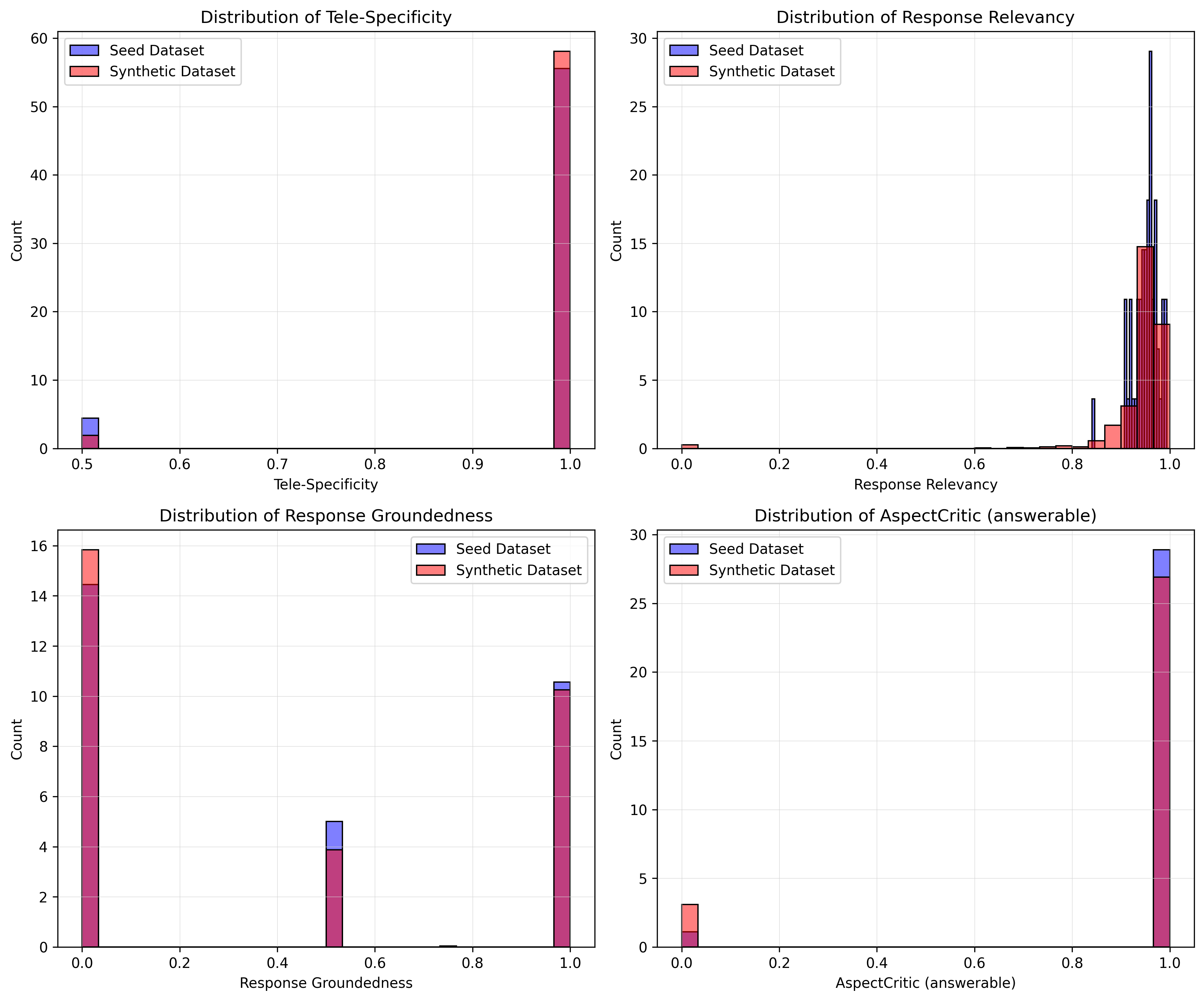}
\caption{\label{fig:metrics_histograms}Distribution of RAGAS metrics between Seed and Synthetic Dataset. Histograms are overlaid with transparency for direct comparison: blue represents seed dataset, red represents synthetic dataset, and overlapping regions appear as a blended color.}
\end{figure}

 To initiate the synthetic data generation process, we curate alarm- and topic-specific inputs from network troubleshooting documents, focusing on 41 topics related to power failure issues. For each topic, the pipeline is configured to generate 10 QA pairs. During question generation, the system randomly samples three examples from the seed dataset as few-shot prompts. Each question is then passed to HippoRAG, which retrieves the top-3 relevant context chunks for the given topic. Leveraging these contexts, Qwen3-14B-Base generates diverse candidate QA pairs. Since base models often struggle with reasoning and producing detailed troubleshooting steps, we further refine the answers with instruct-tuned Qwen3-8B, guided by the retrieved top-3 context chunks from HippoRAG to ensure accuracy and domain grounding. Once sufficient QA pairs are generated, we evaluate the synthetic dataset using customized RAGAS metrics to ensure domain fidelity and quality. Specifically, we establish the following thresholds based on the comparison between the synthetic and seed datasets, as illustrated in Figure~\ref{fig:metrics_histograms}: 

\begin{itemize}
\item Response Groundedness \textgreater\: 0.75
\item AspectCritic (answerable) = 1
\item Tele-Specificity \textgreater\: 0.75
\item Response Relevancy \textgreater\: 0.5
\end{itemize}

\begin{figure}[htbp!]
\centering
\includegraphics[width=1\linewidth]{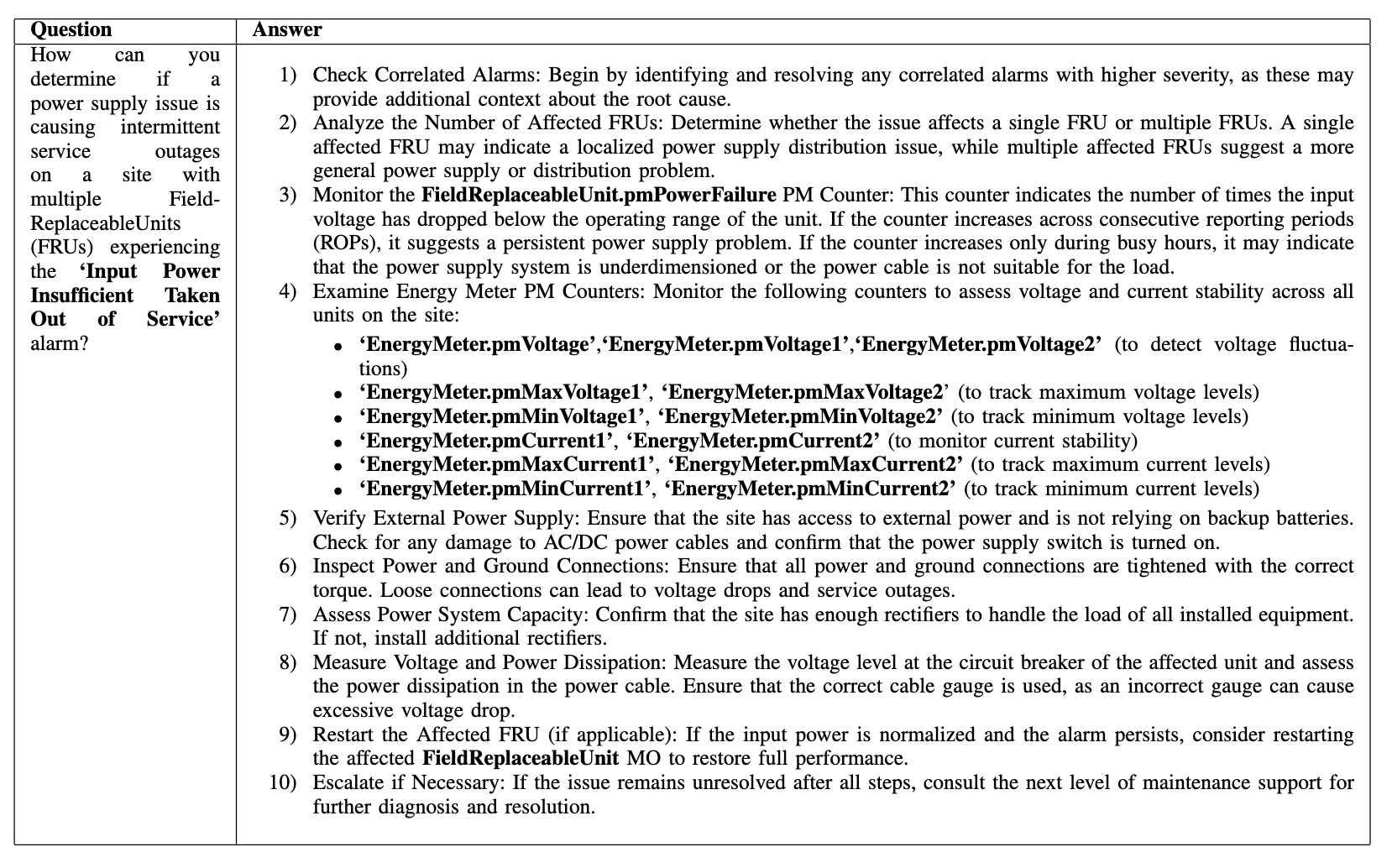}
\caption{\label{fig:RFTsample}Example question–answer pair generated from the synthetic dataset.}
\end{figure}

We set the Response Relevancy threshold to \textgreater{} 0.5 to filter out zero-scoring samples, as the distribution is largely bimodal. Most samples score above 0.8, while a small subset clusters near 0, with virtually no cases in between. We set higher thresholds for other metrics, since they align with retaining synthetic data points that are fully answerable, exhibit strong domain-specific terminology, and are well supported by the retrieved context. These constraints ensure that both the generated questions and answers align with the technical demands of telecom troubleshooting and avoid hallucinations. Due to the discrete nature of groundedness, specificity, and the aspect critic, this ensures that questions will be fully grounded, contain all of the relevant specific details, and will be answerable. Figure~\ref{fig:RFTsample} presents an illustrative QA pair generated from multi-stage domain-grounded for the RFT dataset.

\begin{figure}[htbp!]
\centering
\includegraphics[width=0.6\linewidth]{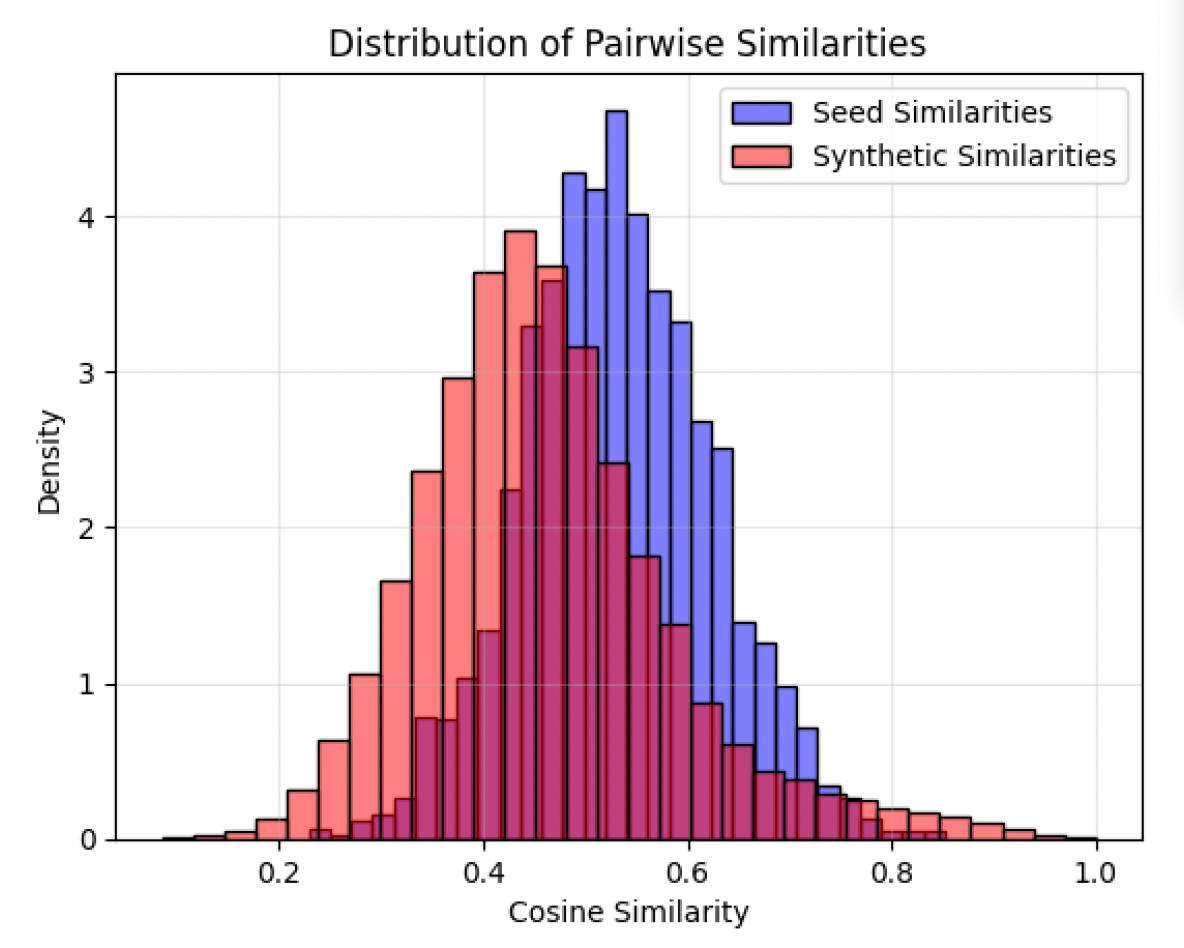}
\caption{\label{fig:BareRagCompare}Distribution of Pairwise Similarities between Seed and Synthetic Dataset. Histograms are overlaid with transparency for direct comparison: blue represents seed similarities, red represents synthetic similarities, and overlapping regions appear as a blended color.}
\end{figure}

\begin{figure}[htbp!]
\centering
\includegraphics[width=0.6\linewidth]{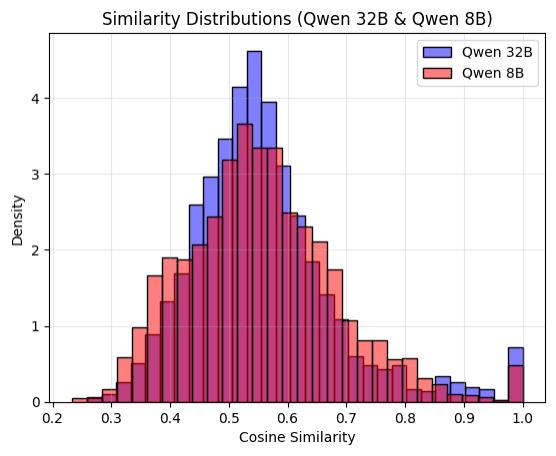}
\caption{\label{fig:BareRagModel}Distribution of Pairwise Similarities between Refiner Qwen3-32B and Qwen3-8B in the Synthetic Dataset. Histograms are overlaid with transparency for direct comparison: blue represents Qwen3-32B, red represents Qwen3-8B, and overlapping regions appear as a blended color.}
\end{figure}

We first evaluate question diversity through computing average neural similarity score \cite{cann2023} by comparing the synthetic dataset with the seed dataset to ensure that the generated questions are sufficiently varied and cover a broader range of topics. Here, we first employ Qwen3-8B as the refiner model. As shown in Figure~\ref{fig:BareRagCompare}, the distribution of pairwise embedding cosine similarity scores for the synthetic dataset follows a similar trend as the seed data but is slightly left-shifted toward lower similarity scores, indicating greater overall diversity. We then switch to a larger LLM, Qwen3-32B, as the refiner model. As shown in Figure~\ref{fig:BareRagModel}, Qwen3-32B exhibits a histogram distribution similar to Qwen3-8B, suggesting that this approach can increase overall diversity of generated data regardless of model size.

\subsection{Comparison of Base-Only, Instruct-tuned-only, and Hybrid Domain-Grounded Synthetic QA Generation}

\begin{figure}[htbp!]
\centering
\includegraphics[width=0.75\linewidth]{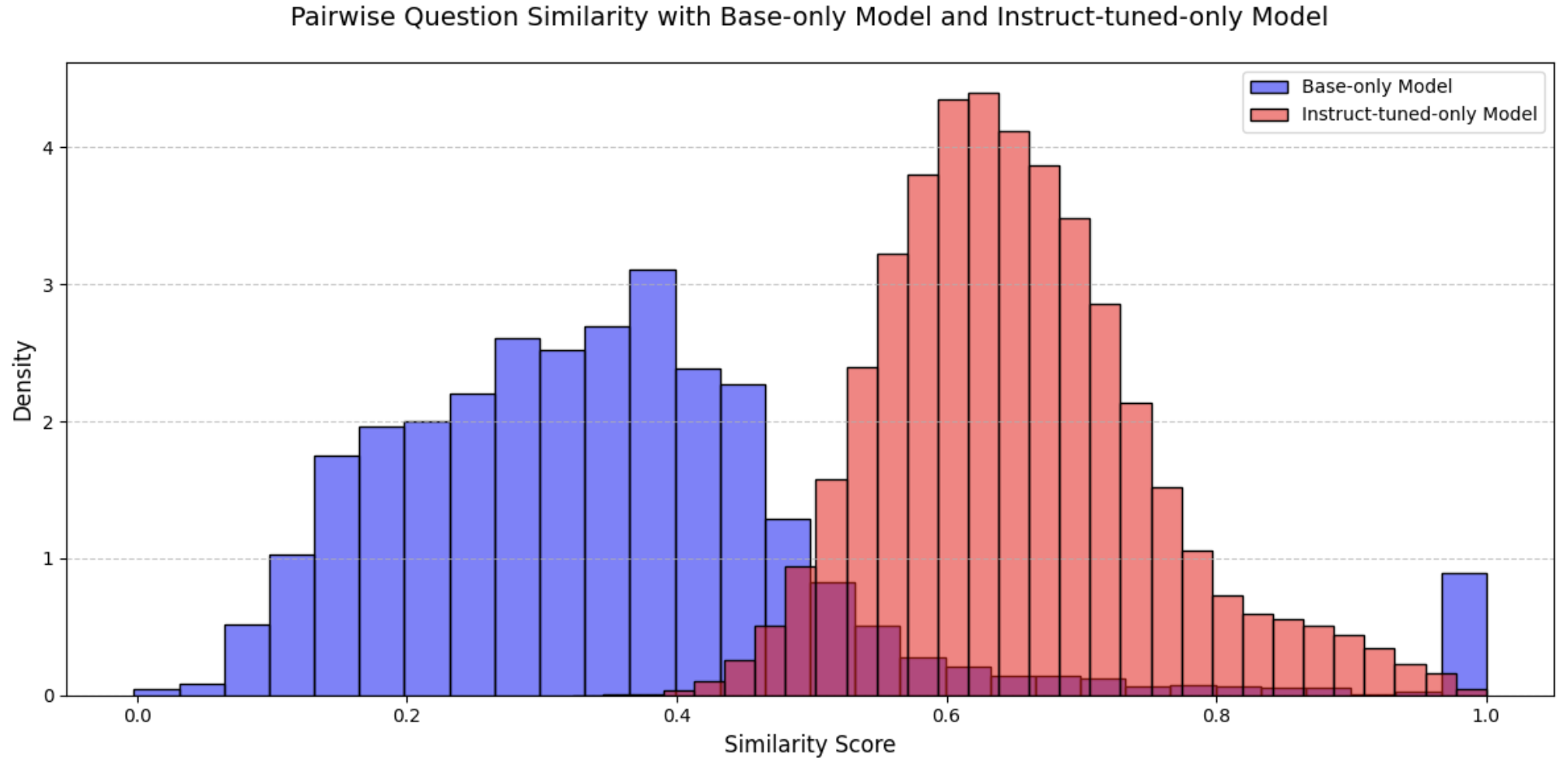}
\caption{\label{fig:BaseAndInstruct}Distribution of Pairwise Similarities with Base-only Model and Instruct-tuned-only Model in the Synthetic Dataset Generation. Histograms are overlaid with transparency for direct comparison: blue represents Base-only model, red represents Instruct-tuned-only model, and overlapping regions appear as a blended color.}
\end{figure}

We compare the hybrid model (Qwen3-14B-Base + Qwen3-8B) against base-only (Qwen3-14B-Base) and instruct-tuned-only (Qwen3-8B) setups in multi-stage domain-grounded synthetic data generation. The hybrid approach not only produces more diverse questions but also delivers detailed, step-wise troubleshooting plans grounded in the top retrieved document chunks after RAGAS evaluation. As shown in Figures~\ref{fig:BaseAndInstruct}, and Table~\ref{tab:SimilarityScore}, the base-only model exhibits higher density in the low-similarity region, which has average similarity score 0.35 compared with 0.65 in instruct-tuned-only, indicating greater question diversity. However, Table \ref{tab:ModelsComparsion} reveals that most base-only outputs fail the RAGAS quality checks, leaving very few usable QA pairs for RFT. While, the hybrid model maintains an appropriate balance between diversity and generation ratio.

\begin{table}[htbp!]
\centering
\caption{Pairwise Question Cosine Similarity Score for BASE-ONLY, INSTRUCT-TUNED-ONLY, AND HYBRID MODEL}
\begin{tabular}{|m{2.6cm}|
                >{\centering\arraybackslash}m{1.5cm}|
                >{\centering\arraybackslash}m{1.5cm}|
                >{\centering\arraybackslash}m{1.5cm}|}
\hline
\textbf{} & \textbf{Base-only Model} & \textbf{Instruct-tuned-only Model} & \textbf{Hybrid Model} \\
\hline
Pairwise Question Cosine Similarity Score & 0.35 & 0.65 & 0.47 \\
\hline
\end{tabular}
\label{tab:SimilarityScore}
\end{table}



\begin{table}[htbp!]
\centering
\caption{Comparison of Base-Only, Instruct-tuned-only, and Hybrid Domain-Grounded Synthetic Troubleshooting QA Generation}
\begin{tabular}{|m{2.6cm}|
                >{\centering\arraybackslash}m{1.5cm}|
                >{\centering\arraybackslash}m{1.5cm}|
                >{\centering\arraybackslash}m{1.5cm}|}
\hline
\textbf{} & \textbf{Base-only Model} & \textbf{Instruct-tuned-only Model} & \textbf{Hybrid Model} \\
\hline
Number of Total Generated QA Pairs & 386 & 394 & 813 \\
\hline
Number of Total QA Pairs after RAGAS Evaluation & 13 & 303 & 273 \\
\hline
Multi-Stage Domain-Grounded Synthetic Data Generation Ratio & 3.4\% & 76.9\% & 33.6\% \\
\hline
\end{tabular}
\label{tab:ModelsComparsion}
\end{table}

We further evaluate entry-wise quality using the indistinguishability rate (IR) \cite{gan}. In this setting, a strong LLM (Qwen3-32B) is employed as a discriminator tasked with identifying a synthetic entry among 3 randomly selected real entries from the seed dataset. The IR measures the rate at which the discriminator fails to distinguish the synthetic entry. A higher IR indicates that the synthetic data closely mimics real data, whereas a lower IR suggests it is easily recognized as out-of-distribution. As shown in Table~\ref{tab:IR}, the instruct-tuned model achieves a higher IR, indicating its stronger ability to generate in-domain examples that closely resemble real-world data. This makes it a practical choice when prioritizing the largest usable dataset with minimal compute cost. However, for large-scale RFT dataset curation where broader coverage and greater diversity are required, the hybrid model becomes more suitable—though it necessitates heavier filtering to ensure quality.

\begin{table}[htbp!]
\centering
\caption{Indistinguishability Rate for Base-only, Instruct-tuned-only, and Hybrid Model}
\begin{tabular}{|m{2.6cm}|
                >{\centering\arraybackslash}m{1.5cm}|
                >{\centering\arraybackslash}m{1.5cm}|
                >{\centering\arraybackslash}m{1.5cm}|}
\hline
\textbf{} & \textbf{Base-only Model} & \textbf{Instruct-tuned-only Model} & \textbf{Hybrid Model} \\
\hline
Indistinguishability Rate (IR) & 69\% & 86\% & 75\% \\
\hline
\end{tabular}
\label{tab:IR}
\end{table}

\subsection{Comparison of Different Chunking on Hybrid Domain-Grounded Synthetic QA Generation}

We evaluate multiple chunking strategies, including semantic and recursive/uniform approaches, applied to the parsed documents to assess their influence on domain-grounded synthetic QA generation. As presented in Table~\ref{tab:ChunksComparsion}, the uniform chunking method yields a notable improvement in the data generation ratio, though this gain is accompanied by a marginal reduction in diversity.

\begin{table}[htbp!]
\centering
\caption{Comparison of different Chunking on Hybrid Domain-Grounded Synthetic QA Generation}
\begin{tabular}{|m{2.6cm}|
                >{\centering\arraybackslash}m{1.5cm}|
                >{\centering\arraybackslash}m{1.5cm}|
                >{\centering\arraybackslash}m{1.5cm}|}
\hline
\textbf{} & \textbf{No Chunking} & \textbf{Semantic Chunking} & \textbf{Uniform Chunking} \\
\hline
Multi-Stage Domain-Grounded Synthetic Data Generation Ratio & 33.6\% & 25.4\% & 49.3\% \\
\hline
Pairwise Question Cosine Similarity Score & 0.47 & 0.45 & 0.51 \\
\hline
\end{tabular}
\label{tab:ChunksComparsion}
\end{table}

\subsection{Runtime Performance}


We optimize runtime performance through several system-level strategies. All Qwen3 models are served via vLLM, with inference managed by LiteLLM and augmented with retry logic to enable robust multi-model routing. For evaluation compatibility, we implement a wrapper that adapts vLLM’s inference engine to the BaseRagasLLM interface. To support long-context reasoning, we enable RoPE scaling in vLLM using the YaRN strategy (factor = 2), allowing efficient processing of the top-3 retrieved context chunks per query. With these optimizations, synthetic data generation completes in approximately 45 minutes, followed by an additional 20 minutes for RAGAS-based quality filtering.

\subsection{Experiments on TeleQuAD Dataset}

We further validate our multi-stage domain-grounded synthetic data generation pipeline on the TeleQuAD dataset \cite{telequad2025}, which contains QA pairs and contexts derived from 3GPP specifications. From these documents, we curate 50 topics and generate 10 examples per topic using three model settings for comparison: base-only, instruct-tuned-only, and hybrid. The same Qwen3 family of models is employed as in earlier experiments. Results, summarized in Table~\ref{tab:ModelsComparsionTeleQuAD}, show that the hybrid model (Qwen3-14B-Base + Qwen3-8B) produces not only more diverse questions but also higher-quality synthetic data, evidenced by lower similarity scores and improved generation ratio and IR.

\begin{table}[htbp!]
\centering
\caption{Comparison of Base-Only, Instruct-tuned-only, and Hybrid Domain-Grounded Synthetic TeleQuAD QA Generation}
\begin{tabular}{|m{2.6cm}|
                >{\centering\arraybackslash}m{1.5cm}|
                >{\centering\arraybackslash}m{1.5cm}|
                >{\centering\arraybackslash}m{1.5cm}|}
\hline
\textbf{} & \textbf{Base-only Model} & \textbf{Instruct-tuned-only Model} & \textbf{Hybrid Model} \\
\hline
Multi-Stage Domain-Grounded Synthetic Data Generation Ratio & 21.5\% & 26.8\% & 49.2\% \\
\hline
Pairwise Question Cosine Similarity Score & 0.36 & 0.60 & 0.36 \\
\hline
Indistinguishability Rate (IR) & 67\% & 88\% & 74\% \\
\hline
\end{tabular}
\label{tab:ModelsComparsionTeleQuAD}
\end{table}

\section{Conclusion}


We present a multi-stage, domain-grounded pipeline for synthetic QA generation to support RFT of LLMs. By integrating HippoRAG retrieval, diverse base generation, structured refinement, and customized RAGAS evaluation, the pipeline produces high-quality, procedurally grounded troubleshooting data without human annotation. Telecom experiments show that the hybrid approach outperforms base-only and instruct-tuned-only baselines, while runtime optimizations enable efficient large-scale generation. Overall, this work offers a scalable path to reducing expert labeling effort and adapting LLMs to knowledge-intensive domains.

\balance

\vspace{-12pt}

\end{document}